# Towards Automated Solar Panel Integrity: Hybrid Deep Feature Extraction for Advanced Surface Defect Identification

**Muhammad Junaid Asif [1*], Muhammad Saad [1,2], Usman Nazakat [1,2], Uzair Khan [1,2], Rana Fayyaz Ahmad [2]**

[1] Artificial Intelligence Technology Centre (AITeC), National Centre for Physics (NCP), Islamabad 44000, Pakistan
[2] School of Computing, Pak-Austria Fachhochschule: Institute of Applied Sciences and Technology (PAFIAST), Haripur 22620, Pakistan

*e-mail: junaid.asif@ncp.edu.pk

*Abstract*— **The monitoring of PV panels is crucial for energy efficiency and optimizing power production. Manual inspection of large scale and out-of-the-way solar plants, though, is labor intensive, time consuming, expensive and can be prone to human error. This research introduces an automated mixed feature-based approach for PV panel defect classification using both handcrafted and deep feature representations. Firstly, image preprocessing and augmentation are used to enhance image quality and diversify the data. The handcrafted representations are learned from Local Binary Patterns (LBP), Histogram of Oriented Gradients (HOG) and Gabor filters, while the deep representations are learned using DenseNet-169. Various individual and fused feature configurations are then tested by Support Vector Machine (SVM), XGBoost and LightGBM classifiers. Experimental evidence shows that representations that are deep and selectively fused provide significant improvement over stand-alone handcrafted representations. As for the handcrafted methods, LBP + LightGBM method achieved accuracy of 91.85% and dual handcrafted fusion resulted in an accuracy of 94.23% on LightGBM on handcrafted features (LBP + Gabor). DenseNet-169 with SVM had a recognition accuracy of 99.17% while the DenseNet-169 + Gabor with SVM hybrid achieved the accuracy of 99.17% also. However, the fusion of multiple features that were extensive, such as triple and quadruple feature fusion, led to decreased performance, suggesting that the more the dimensionality and redundancy of features, the lower the classification performance. These results show that selective fusion of complementary handcrafted and deep features is an effective method for automated classification of PV panel defects and could be used for intelligent PV panel monitoring applications.**

## I. Introduction

Electricity is a vital energy source that underpins residential, industrial, and technological applications. It can be generated both by conventional and renewable sources. Oil, coal, nuclear, and natural gas are among the transitional sources [1], and are linked to severe health and ecological problems because they release hazardous pollutants and industrial waste. Alternatively, renewable energy sources such as SOLAR, wind, hydropower and geothermal are healthier, more resilient, and have a considerably lower environmental impact [2]. To reach global sustainability and boost quality of life, it is crucial to make better use of energy resources by using renewable energy resources [3]. The trend to renewable electricity generation has been speeded up by rising environmental, health and societal concerns. The major renewable energy sources are: solar, wind, biomass, hydropower and geothermal.

Solar energy is one of the cleanest renewable energy sources, generating electricity when light strikes a solar cell without creating harmful emissions [4]. The flexibility that it provides allows for its deployment on both residential rooftops and on larger scale installations and is especially useful for off-grid areas [5]. PV panels, however, suffer from degradation over time which will cause a reduction in efficiency and highlight the importance of effective fault detection and monitoring [6]. Furthermore, solar energy generation is also influenced by the weather, which can impact solar energy production, for example, through rainfall and cloud and precipitation cover [7].

Solar panels may also face an array of challenges that make them less efficient at generating electricity. Some common visual problems ***(as shown in Fig. 1)*** include dust, bird droppings, snow cover, electrical faults, and physical damage. Dust particles may accumulate on the surface of solar plates, making it harder for sunlight to reach them and reducing the generation capacity of electrical power. The same thing happens when bird droppings build up: they block light from entering. In colder areas, snow may accumulate on solar panels for extended periods, making it even harder to generate energy.

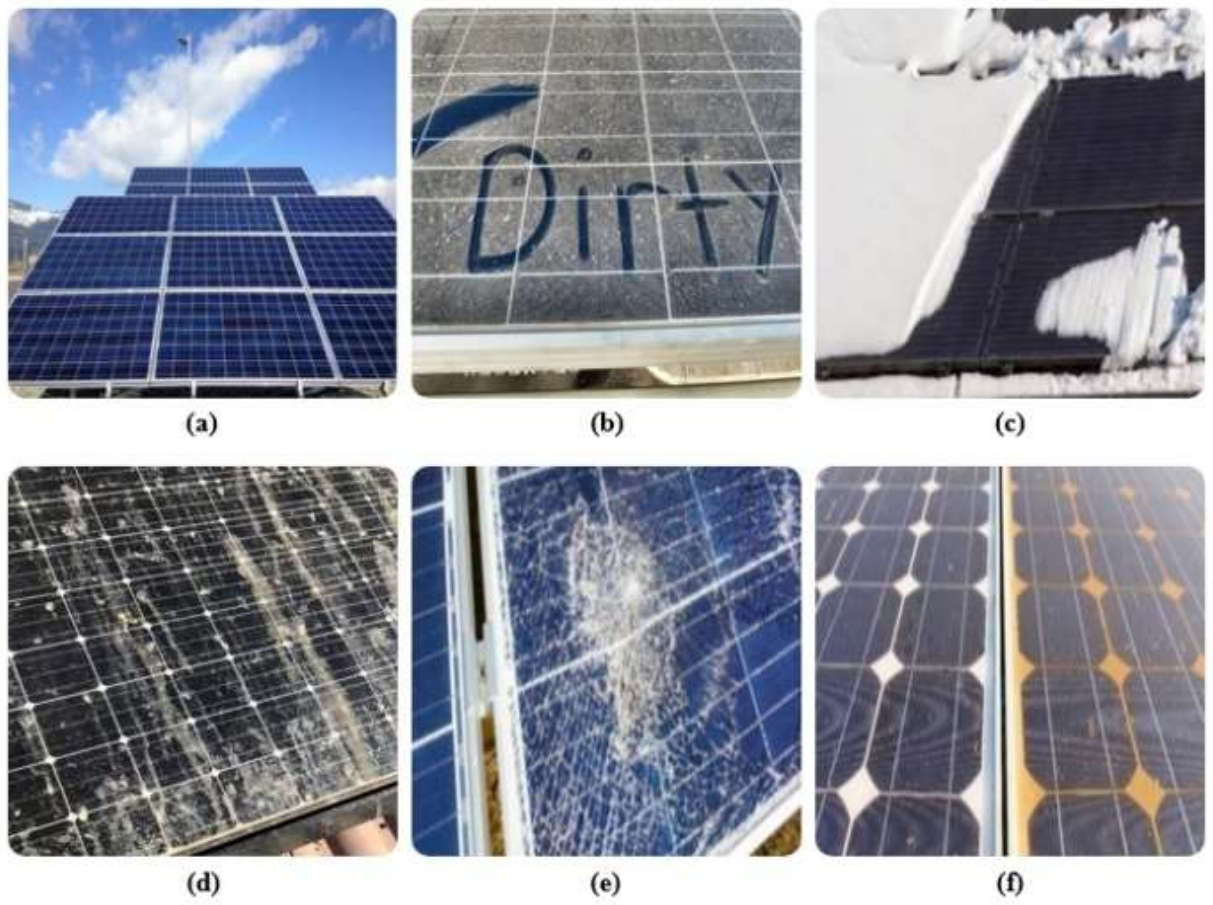

**Fig. 1.** Visual Samples Depicting Various Surface Conditions of PV Panels for Defect Detection: (a) Clean, (b) Dust, (c) Snow-covered, (d) Bird Droppings, (e) Physical Damage, and (f) Electrical Fault

External factors, including hail hitting the solar glass or cells, can cause physical damage, such as cracks. Overheating may result in electrical issues, leading to burnout of solar cells overheating of solar panels leading to the burn out of solar cells [8]. The quality inspection processes heavily rely on defect detection. **Defect or Anomaly detection (AD)** process aims at detecting abnormal features that may lead to quality or functionality issues of a particular object. As solar installations continue to expand globally, it has become essential to ensure they operate effectively and last a long time. Regular physical inspection is one of the most essential steps to maintain the quality and reliability of solar plates. It is quite a challenging and laborious task to perform the manual inspection of large-scale solar plants installed at governmental or industrial scale. Manual inspection is becoming impractical as the number of solar panel installations continues to increase [9].

There is a pressing need to propose an accurate and efficient method for automatic inspection of PV panels to ensure their long operational life span and maintain high energy efficiency. Due to recent advancements in the field of artificial intelligence, computer vision and deep learning, researchers from different research domains are entering in the areas of automated defect detection such as surface crack detection [10], product defect detection [11] , and damage detection. To address the challenges associated with manual inspection, we proposed a hybrid framework combining both deep and machine learning techniques for efficient and accurate inspection of solar plates. The proposed framework consists of four major components: data augmentation, data preprocessing, **Feature Extraction Network (FEN)** and defect classification. A publicly available dataset on KAGGLE, consisting of **1,574 images**, is used for experimentation purposes. Due to the low volume of the dataset and to reduce the risk of overfitting, three different data augmentation operations, such as rotation, flipping, and image translation were applied. This overall augmentation process resulted in a fourfold increase in dataset size, reaching about **6,060 images.**

Different image preprocessing techniques were applied to reduce noise and improve image quality. A hybrid feature extraction framework was then employed, combining handcrafted descriptors—**LBP [12], HOG [13], and Gabor filters [14]**—with deep features extracted using **CNNs [15].** The handcrafted and deep features were concatenated into a unified feature representation and subsequently classified using machine learning classifiers, including **SVM [16], ANN[17], LR [18], and XGBoost [19],** for accurate PV panel defect classification. The overall architecture of the proposed framework is illustrated in **Fig. 2.**

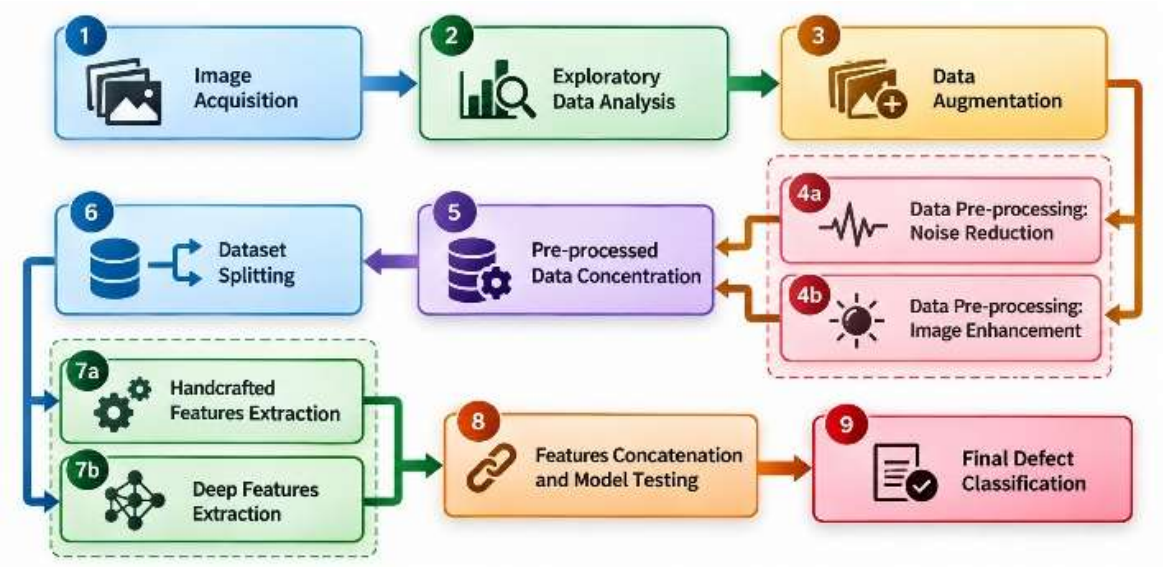

**Fig. 2.** Block diagram illustrates the overall architecture of the proposed system for PV panel defect detection.

The contributions of this research are as follows:

- An optimized data augmentation pipeline is proposed, resulting in a fourfold increase in data volume to improve model generalization.
- Proposed an automatic, accurate and efficient inspection system based on a hybrid network that combines handcrafted features (extracted using LBP, HoG and Gabor) with deep learning features extracted by leveraging the use of CNN to enhance the classification accuracy.
- The novel feature-level fusion method combines both handcrafted and deep features through a concatenation into a single feature matrix, to obtain optimal classification by combining the limited strengths of other features.
- Fused features were tested with various machine learning classifiers (SVM, XGBoost, ANN, Logistic Regression) and comparable results showed improved performance of fused features compared to current single feature models.

The paper will be structured as follows: **Section II** will provide an overview of recent methodologies applied for detecting defects in SOLAR panels. **Section III** outlines the proposed methodology, which includes a description of the dataset used, data processing, and hybrid feature extraction. **Section IV** describes the experimental design, presents the qualitative and quantitative findings, and offers a comparison with previous methods. **Section V** discusses the relevance of the findings and their implications to real-world use of PV monitoring. Lastly, **Section VI** concludes the paper by summarizing the main contributions and outlining possible future research directions.

## II. Related Work

### *2.1. Traditional Approaches:*

Earlier approaches to product defect detection primarily focused on traditional image processing techniques that generally concentrated on capturing handcrafted features such as geometric structure, color, shape, edge patterns, and surface roughness or intensity variations, using traditional feature descriptors or statistical techniques. For instance, Tsai et al. [20] used Fourier image reconstruction to detect defects, such as cracks, breaks, and interruptions, in electroluminescence (EL) images of polycrystalline solar modules. They used their methodology of filtering the high-frequency components in the Fourier domain and identifying defects comparing the intensity of the original and reconstructed images. The method, however, demonstrated lower effectiveness when subjected to the defects of a difficult nature. Author proposed another method for solar plate defect-detection based on Independent Component Analysis (ICA) by optimizing the feature extraction and reconstruction steps and making them much more effective in slight variations in texture and intensity in EL images. On the same note, Zhang et al. [21] came up with an Independent Component Analysis (ICA)-based reconstruction algorithm, where a defect-free image was used as a reference to help in the extraction of independent components that would be used in defect detection. The reconstructed images were very effective at highlighting local anomalies; hence, the technique was appropriate for revealing local defects on textured surfaces with periodic features. Both Fourier image reconstruction and ICA-based methods have severe limitations in handling complex, irregular shaped defects. The Fourier-based method is sensitive to noise and light variations, thus limiting its accuracy across diverse environmental conditions. Similarly, ICA also depends on high-quality reference images and experiences challenges with periodic or less visible anomalies, thus limiting its practical usefulness for inspecting solar panels.

### *2.2. Deep Learning Approaches:*

Due to the challenges faced by traditional image processing and machine learning approaches, deep learning approaches have emerged as state-of-the-art techniques for tasks related to image analysis. In this regard, Convolutional neural networks CNNs) shows promising results by capturing both local as well as global features for identification and classification of defects. Shaik et al. [22] proposed a novel approach to enhance the monitoring and defect-detection for solar panels, addressing the challenges of traditional manual inspections, particularly at scale. By utilizing Unmanned Aerial Vehicles (UAVs) and satellite imagery, a more rapid and efficient monitoring system has been established. A deep learning-based segmentation U-NET model integrated with an **Atrous Spatial Pyramid Pooling (ASPP) module** was proposed, achieving an impressive overall accuracy of 98% and a Mean Intersection-Over-Union (IoU) Score of 95%. The proposed system efficiently detects five classes of damage, including undamaged solar panels, demonstrating high level of effectiveness in defect identification.

Another author [23] proposed a deep learning-based framework that utilizes the thermal images for the identification and classification of defects in PV plates. The proposed model, based on Convolutional Neural Network with VGG16 backbone, achieves an impressive F1-Score of 84.12% and Mean Minimum Precision (MMP) of 71%, which may be explained by using localization heatmaps to identify errors with high precision. Furthermore, the model's adaptability and strong competitiveness were demonstrated through a comparative analysis with advanced models including Yolo-v8 and transformer-based architectures such as Vision Transformers (ViT) and Swin Transformers. Both techniques, as suggested[23], [22] rely heavily on modalities of images such as EL or thermal imaging and hence, their usability in a variety of environmental situations is limited. UAV- and satellite-based systems, although efficient, face the problem of scaling, which is driven by high computational and operational costs. A framework based on MobileNet-V3 was proposed by Mansurov et al.[24], achieving 100% accuracy in identifying and classifying PV panel defects, such as bird-dropping and snow cover. Another author, Kazemi et al. [25] suggested using transfer learning with a pretrained VGG-16 model as a backbone and achieved an accuracy of 74%. These findings indicate remarkable progress toward the automatic diagnosis of faults in solar panels, opening a new horizon for energy efficiency and fewer manual inspections for privacy. DenseNet-121, also suggested by another author [26], achieves an accuracy of 86.44%.

Nagar et al. [27] explored the DenseNet169 model using a dataset of 11,000 images across six classes: bird droppings, dust, electrical damage, physical damage, clean panels, and snow-covered panels. Impressively, the DenseNet169 model achieved high training accuracy and demonstrated robust performance on the test dataset, demonstrating its powers in enhancing the efficiency of solar panel inspection and maintenance procedures. Numerous deep learning algorithms as suggested by [24], [25], [26] including VGG16, DenseNet and MobileNetV3 requires large, annotated data sets and are not generalized to new defect types or real-world changes in brightness, particulate and specular reflection. In addition, models that have been trained using constrained data tend to perform poorly under diverse field conditions. Finally, privacy-preserving strategies, such as federated learning, are frequently compromised for detection accuracy, indicating the need for more balanced and dynamic models.

Recent research has made a valuable contribution to the methodology for detecting and classifying the defects in solar panels; however, several critical issues remain. The existing literature is limited primarily to focusing on a limited list of types of defects. It is based on relatively limited volume of images, typically fewer than 1,500 samples, thus limiting the ability of the models to apply to a wide range of environmental conditions. Even though network architectures like VGG16 and YOLO have shown significant promise in the deep-learning paradigm, their performance is highly dependent on large, labelled data sets and they often struggle with real-world variability. In addition, the potential advantages of hybrid models combining handcrafted feature engineering with deep-learning models are not studied thoroughly, which ends up in poor accuracy and robustness. To address these shortcomings, the current paper proposes a hybrid model, that combines handcrafted and deep feature extraction methods and is further supported by extensive data augmentation. The proposed strategy will increase datasets, thereby improving the effectiveness and accuracy of defect detection systems.

## III. MATERIALS AND METHODS

### 3.1. *Data Acquisition:*

The **PV Panel Defect Dataset** is publicly available on KAGGLE, generally designed and developed for detection and classification of different faults in photovoltaic (PV) panels. It contains six classes representing real-world conditions for solar panels, including clean, dusty, bird-dropping, electrical damage, physical damage and snow-covered panels. The dataset contains high-resolution RGB photographs taken under various lighting and environmental conditions, ensuring the resilience and generalizability of deep learning and computer vision-based fault detection systems. **Table 1** presents an overview of the PV Panel Defect Dataset.

Table 1. PV PANEL DEFECT DATASET SUMMARY

| Class | Description | No. of Images |
|---|---|---|
| **Clean** | Images of clean solar panels. | 289 |
| **Snow-covered** | Images of solar panels covered with snow. | 262 |
| **Dusty** | Images of solar panels covered with dust | 275 |
| **Electrical Fault** | Images of solar panels damaged due to electrical fault. | 225 |
| **Physical Damage** | Images of physically damaged solar panels | 225 |
| **Bird Droppings** | Images of solar panels covered with bird droppings | 298 |

### 3.2. *Data Preprocessing:*

#### 3.2.1. *Noise Reduction*

The methods employed to improve the image quality are **Non-Local Means (NLM)** denoising and Bilateral filtering. Non-Local Means (NLM) denoising was effective at reducing noise without losing important textural and structural features needed to detect defects accurately, whereas **Bilateral filtering** improved image smoothness while preserving important edge details. A bilateral filter with a ***diameter of 9, sigmaColor of 75, and sigmaSpace of 75*** was applied, which was sufficient to reduce image noise while retaining the important features such as edges. NLM denoising was also applied with the following parameters: ***h=10, hColor=10, the size of the template window = 7 and the size of the search window = 21.*** This overall denoising operation helped successfully remove noise without damaging any important textural and structural information of the image.

### 3.2.2. *Image Enhancement*

Two different image enhancement techniques were employed to improve contrast and brightness consistency, including ***Contrast Limited Adaptive Histogram Equalization (CLAHE)*** and Gamma Correction. Firstly, the images were downsized to **640x640 pixels**, followed by converting the BGR to the LAB to increase luminance while preserving chromaticity. The L channel was then processed with **Contrast-Limited Adaptive Histogram Equalizations (CLAHE)** with a **clip limit of 2.0** and a **tile grid of 8×8**. Lastly, the equalized L channel was reassembled with the respective **A and B channels** and then lastly converted to the BGR color model once again, giving the images a brighter contrast and clarity improving the ability to detect defects. In addition, gamma correction was applied to improve image contrast. It is a nonlinear enhancement method that changes the value of pixel intensities according to power-law function. in accordance with a power-law function. The gamma value in our research was 1.5, used to lighten mid-tone areas without altering the overall image structure. A lookup table was built to effectively transform the original pixel values into gamma adjusted ones. All the images were resampled to 640x640 pixel size and subjected to this transformation, yielding a collection of contrast-enhanced images in a different direction.

### 3.3. *Feature Extraction Network:*

In our research work, we proposed a hybrid feature extraction model based on combination of handcrafted feature extraction techniques and deep learning algorithms to take advantage of both approaches. This combination will enable the model to capture features simultaneously both at low level (textual and edge-related data) and high level (semantic)., resulting in development of an accurate and efficient defect detection system.

### 3.3.1. *Handcrafted Features*

For handcrafted features extraction, we employed three different approaches: Linear Binary Patterns (LBP), Histogram of Gradients (HoG) and Gabor Filters. LBP is applied for textural analysis, HoG is used for capturing shape and edges, and Gabor filters are used for analyzing the spatial frequency and orientation. The combination of these methods can provide an extensive overview of surface aspects, thus enhancing overall classification accuracy.

### 3.3.2. *Deep Learning Features*

For deep features extraction, we used Convolutional Neural Networks (CNNs) with backbone of DenseNet169, that provides high level semantic features. This helped us to extract complex and intricate features that couldn't be captured by handcrafted approaches.

### 3.3.3. *Feature Fusion*

To leverage the use of both handcrafted features and deep features, both were concatenated in the form of single feature matrix, which integrated low-level visual characteristics with high-level abstractions.

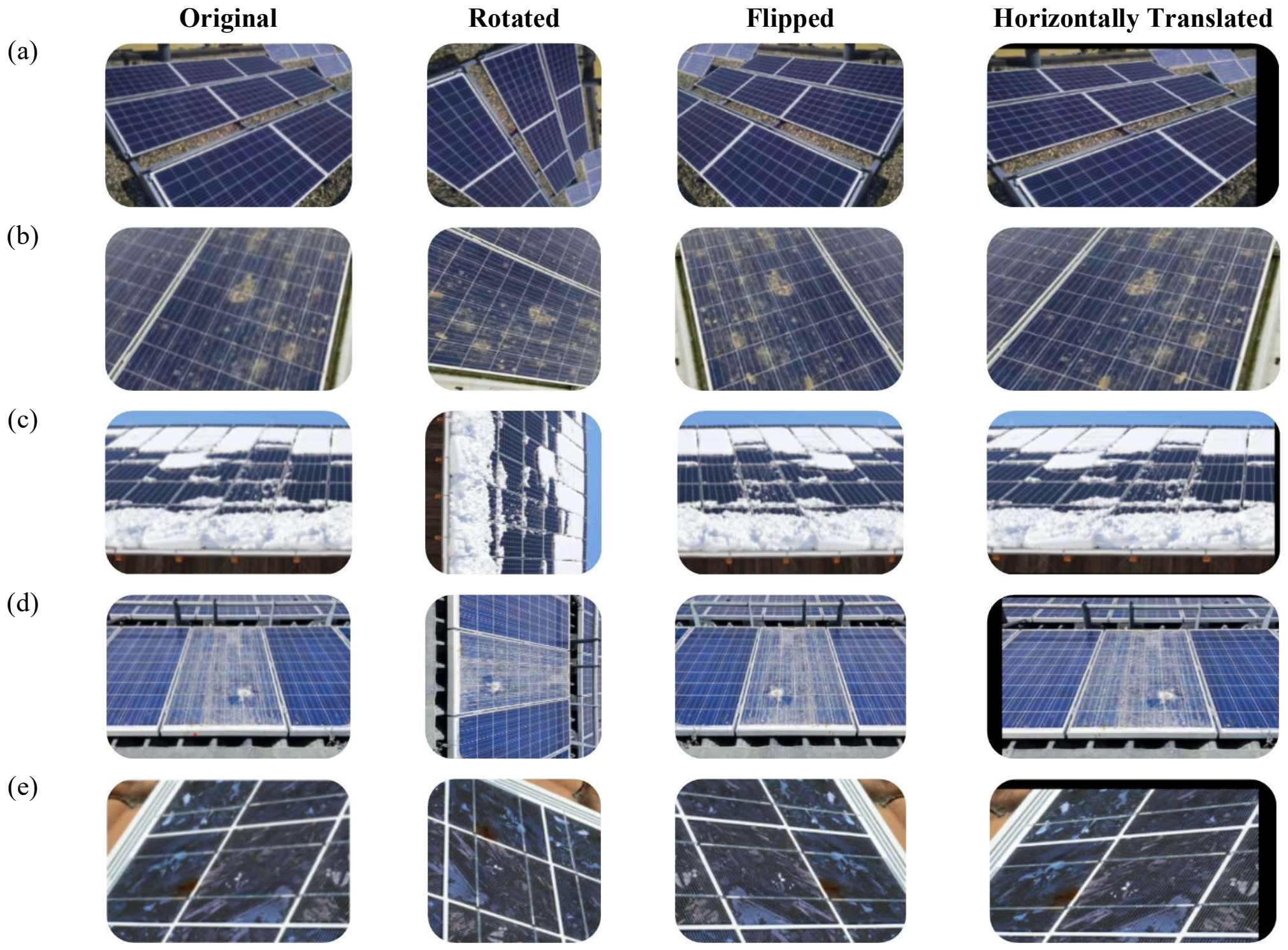

**Fig. 3.** Illustration of data augmentation techniques applied to PV panel images. (a) Clean panels, (b) Panels affected by bird droppings, (c) Snow-covered panels, (d) Panels with physical defects, and (e) Panels exhibiting electrical faults.

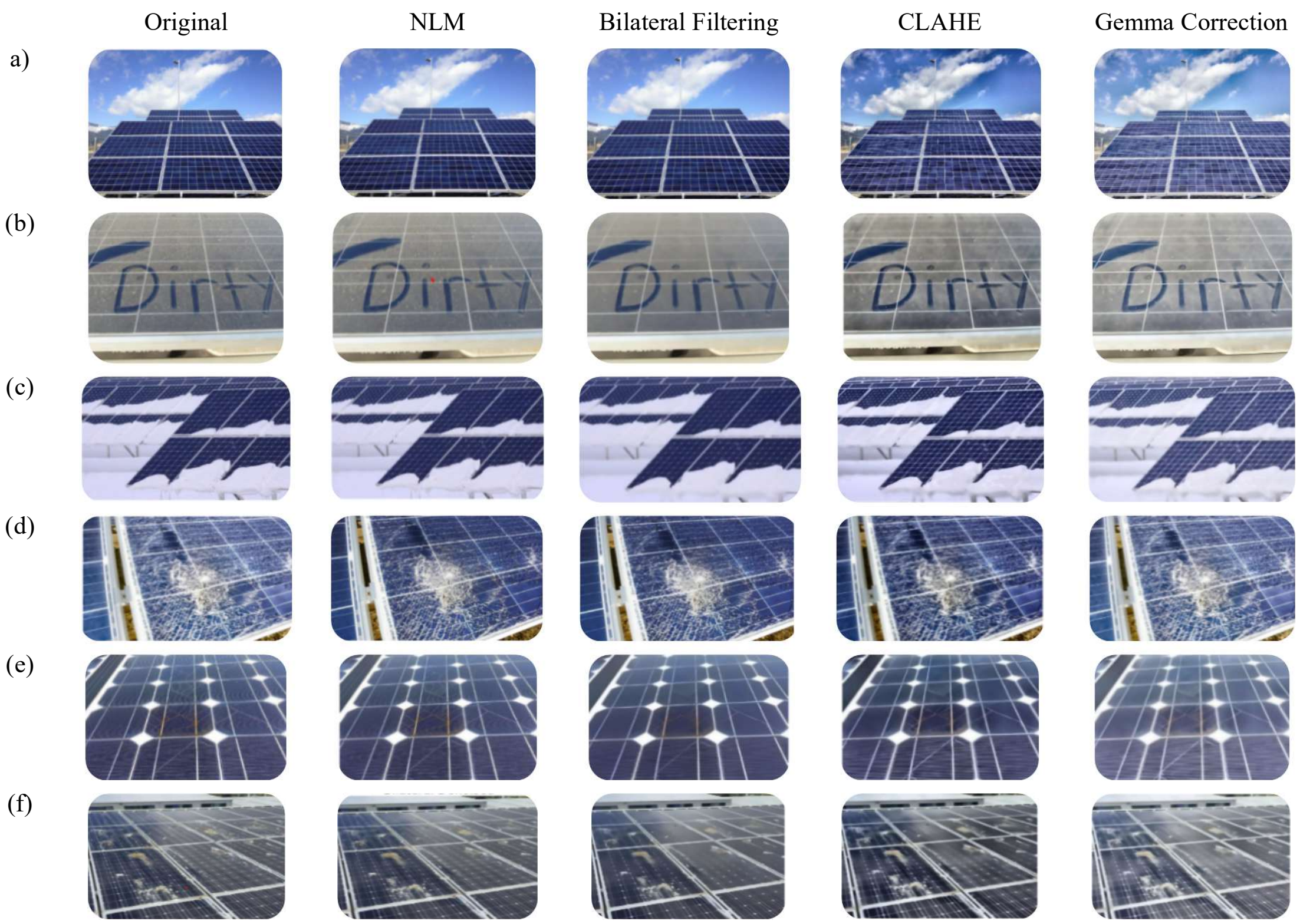

**Fig. 4.** Illustration of preprocessing techniques applied to PV panel images. (a) Clean panels, (b) Panels covered with Dust, (c) Snow-covered panels, (d) Panels with physical defects, and (e) Panels exhibiting electrical faults, (f). Panels affected by bird droppings.

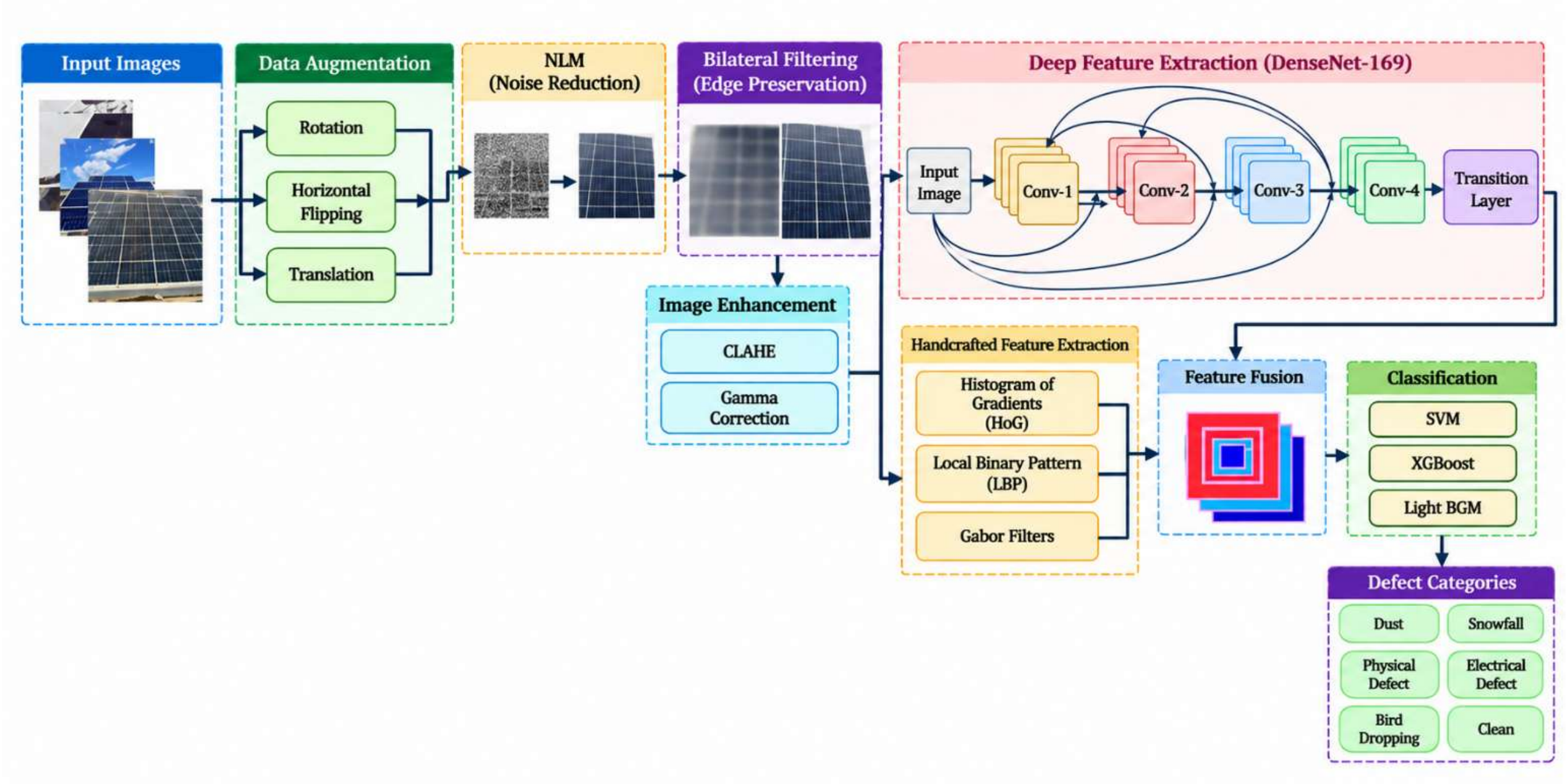

**Fig. 5.** Overall architecture of the proposed hybrid PV panel defect detection framework.

### 3.4. *Classification:*

The fused features are then fed to classification network, comprising three different classifiers: Support Vector Machines (SVMs), Extreme Gradient Boosting (XGBoost) and Light Gradient Boosting Machine (Light GBM). Each of the ML classifiers are selected due to its unique strength in handling complex and high-dimensional.

We employed SVM because of its excellent ability to build optimal decision boundaries between the classes by the application of kernel-based optimization. It is well suited for small-to-medium size datasets, and it is stronger at high classification accuracy even with restricted training samples, thus suitably applicable to PV defect classification where inter-class distances may be very fine.

*3.5. Evaluation Metrics:*

Performance of our proposed model was being evaluated using different evaluation metrics such as accuracy, precision, recall and F1-score. These metrics were selected to provide a comprehensive assessment of the model's classification capability. Accuracy is used to determine the general validity of predictions, whereas precision is used to measure the percentage of correct samples of defects out of all predicted defects. Recall is used to test the sensitivity of the model (its ability to detect actual defects) and F1-score (harmonic-mean between precision and recall) is a balanced measure of model performance [28], [29], [30].

$$\text{Accuracy} = \frac{TP+TN}{TP+TN+FP} \quad (1)$$

$$\text{Precision} = \frac{TP}{TP+F} \quad (2)$$

$$\text{Recall} = \frac{TP}{TP+FN} \quad (3)$$

$$F1-\text{Score} = 2 \times \frac{\text{Precision x Recall}}{\text{Precision+}} \quad (4)$$

## IV. Results and Discussions

*4.1. Performance of Handcrafted Feature Extraction Techniques:*

**Table 2** shows the performance of three handcrafted feature extraction techniques, along with **SVM, XGBoost and LightGBM (LGBM)** classifiers. **LBP + LGBM** achieved the highest accuracy of **91.85%,** which showed that the LBP method has a good performance in the extraction of discriminative local texture patterns and the evaluation of PV panel condition including dust and snow. However, the HOG based approaches had relatively poorer accuracy of ~80-85%, indicating that gradient based shape and edge information may not be enough to distinguish between the subtle differences in surface between defect classes. The Gabor-based models gave moderate accuracy of around 86-90% and proved to be effective in capturing spatial-frequency and orientation information but may be affected by changes in illumination and/or image texture.

Table 2. Comparative Performance Analysis of Handcrafted Feature Extraction Techniques Combined with Various Classifiers

| Feature Extraction Method(s) | Classifier | Accuracy (%) | Precision (%) | Recall (%) | F1-Score (%) |
|---|---|---|---|---|---|
| Gabor | SVM | 86.50 | 86.83 | 86.60 | 86.60 |
| Gabor | XGBoost | 89.50 | 89.80 | 99.60 | 89.50 |
| Gabor | LGBM | 89.38 | 90.01 | 89.30 | 89.36 |
| HOG | SVM | 84.77 | 84.99 | 84.77 | 84.80 |
| HOG | XGBoost | 79.42 | 81.01 | 79.42 | 79.44 |
| HOG | LGBM | 81.97 | 82.32 | 81.97 | 81.94 |
| LBP | SVM | 90.01 | 91.16 | 90.81 | 90.81 |
| LBP | XGBoost | 91.35 | 91.52 | 91.85 | 91.84 |
| **LBP** | **LGBM** | **91.85** | **92.01** | **91.85** | **91.84** |

*4.2. Performance of Dual Handcrafted Feature Extraction Techniques*

As presented in **Table 3**, the combination of two handcrafted feature descriptors proved useful to build classification system, which incorporates complementary visual information. The highest accuracy, **94.23%**, was obtained by the **LBP + Gabor** combination with **LightGBM**. This improvement is due to the complementary nature of both LBP and Gabor features, as the former features are used to capture local texture patterns while the latter features are used to capture spatial-frequency information, facilitating the discrimination of different PV panel conditions. LBP + HOG and HOG + Gabor, however, obtained comparatively moderate results and showed that the fusion of features does not necessarily lead to better results. The complementarity of the features is an important factor for the effectiveness of fusion, and redundant or highly correlated features may lead to a high feature dimensionality without adding discriminative information.

Table 3. Comparative Performance Analysis of Dual Handcrafted Feature Extraction Techniques Combined with Various Classifiers

| Feature Extraction Method (s) | Classifier | Acc. (%) | Precis. (%) | Recall (%) | F1-Score (%) |
|---|---|---|---|---|---|
| LBP + HOG | SVM | 85.10 | 85.83 | 85.66 | 85.66 |
| LBP + HOG | XGBoost | 83.70 | 84.50 | 84.33 | 84.16 |
| LBP + HOG | LGBM | 88.14 | 89.01 | 88.14 | 88.08 |
| HOG + Gabor | SVM | 84.60 | 85.33 | 85.01 | 85.01 |
| HOG + Gabor | XGBoost | 84.03 | 84.50 | 84.16 | 84.16 |
| HOG + Gabor | LGBM | 86.01 | 85.60 | 85.30 | 86.01 |
| LBP + Gabor | SVM | 91.17 | 92.01 | 91.76 | 91.76 |
| LBP + Gabor | XGBoost | 91.44 | 91.33 | 91.44 | 91.42 |
| **LBP + Gabor** | **LGBM** | **94.23** | **94.33** | **94.23** | **94.42** |

*4.3. Performance of Tripple Handcrafted Feature Extraction Techniques:*

The combined model of **three handcrafted features (LBP, HOG, and Gabor)** yielded the highest accuracy of **88.80%** using **LightGBM model**, which was less than the model with the best accuracy of the combination of two handcrafted features shown in Table 4. The reduction in performance could be explained by the fact that redundant features and higher dimensional data cause the fused representation to contain redundant or overlapped information, thereby decreasing the power of the discriminative representation. The use of several descriptors will provide more feature space, but, if feature fusion is used without suitable feature selection or dimensionality reduction, it may introduce the danger of overfitting, especially in relatively small datasets.

Table 4. Comparative Performance Analysis of Triple Handcrafted Feature Extraction Techniques Combined with Various Classifiers

| Feature Extraction Method (s) | Classifier | Acc. (%) | Precis. (%) | Recall (%) | F1-Score (%) |
|---|---|---|---|---|---|
| LBP + HOG + Gabor | SVM | 85.43 | 86.16 | 85.43 | 85.42 |
| LBP + HOG + Gabor | XGBoost | 88.39 | 88.66 | 88.39 | 88.34 |
| **LBP + HOG + Gabor** | **LGBM** | **88.80** | **89.01** | **88.87** | **88.01** |

*4.4. Performance of Deep Learning Feature Extraction Techniques:*

The DenseNet-169 (DN-169) model significantly outperformed the manual feature-based models as shown in

Table 5. The strong discriminative power of deep hierarchical features for both local and global visual patterns is illustrated in **DenseNet-169** with the highest accuracy of **99.17%** in combination with **SVM classifier**. The exceptional performance demonstrates the success of deep representations in the detection of complex surface defects in PV. But deep networks like DenseNet-169 are expensive to train and typically need ample and diverse training data to learn well.

Table 5. COMPARATIVE PERFORMANCE ANALYSIS OF DEEP LEARNING FEATURE EXTRACTION TECHNIQUES COMBINED WITH VARIOUS CLASSIFIERS

| Feature Extraction Method (s) | Classifiers | Accuracy (%) | Precision (%) | Recall (%) | F1-Score (%) |
|---|---|---|---|---|---|
| DN-169 | XGBoost | 97.03 | 97.01 | 97.03 | 97.02 |
| DN-169 | LGBM | 97.94 | 98.16 | 97.94 | 97.93 |
| **DN-169** | **SVM** | **99.17** | **99.01** | **99.10** | **99.01** |

### *4.5. Performance of Hybrid Feature Extraction Techniques:*

A hybrid feature extraction system was designed to fuse the benefits of handcrafted features and deep features. DenseNet-169 extracted features were combined with LBP features, HOG features, and Gabor features and classified with SVM, XGBoost and LightGBM classifiers. The **DenseNet-169 + Gabor with SVM configuration** yielded the highest accuracy of **99.17%,** wherein high-level semantic information is used with texture and spatial-frequency features, showing that the two can be used together to obtain good results. DenseNet-169 + LBP and DenseNet-169 + HOG combinations also outperform 98%, showing that handcrafted descriptors can be an effective complement to deep features and effective for robust PV defect classification.

Table 6. COMPARATIVE PERFORMANCE ANALYSIS OF DUAL HYBRID FEATURE EXTRACTION TECHNIQUES COMBINED WITH VARIOUS CLASSIFIERS

| Feature Extraction Method (s) | Classifiers | Acc. (%) | Precis. (%) | Recall (%) | F1-Score (%) |
|---|---|---|---|---|---|
| DN-169 + LBP | SVM | 99.01 | 99.00 | 98.01 | 99.01 |
| DN-169 + LBP | XGBoost | 97.01 | 98.00 | 98.00 | 98.00 |
| DN-169 + LBP | LGBM | 98.10 | 98.01 | 98.10 | 98.10 |
| **DN-169 + Gabor** | **SVM** | **99.17** | **99.16** | **99.01** | **99.17** |
| DN-169 + Gabor | XGBoost | 97.69 | 97.66 | 97.55 | 97.66 |
| DN-169 + Gabor | LGBM | 97.94 | 98.01 | 97.94 | 97.93 |
| DN-169 + HOG | SVM | 92.75 | 93.01 | 92.70 | 92.75 |
| DN-169 + HOG | XGBoost | 97.28 | 97.01 | 97.28 | 97.27 |
| DN-169 + HOG | LGBM | 98.18 | 98.33 | 98.17 | 98.18 |

Finally, as detailed in **Tables 7 and 8**, DenseNet-169 is integrated with several handcrafted descriptors such as LBP, HOG, and Gabor and tested with SVM, XGBoost, and LightGBM. The LightGBM model yielded the highest accuracy of 92.88%, whereas the other models had similar performance in comparison. The reduction in comparison with the simpler configurations is due to feature redundancy and/or higher dimensionality, where several descriptors can include information that is redundant or less discriminative. Furthermore, it may lead to an increase in the risk of overfitting, noise propagation especially when the data is small, and increase the computational complexity. This shows that not all additional features are statistically significant in improving classification performance and highlights the need for feature selection or dimensionality reduction in the construction of high dimensional hybrid representations.

Table 7. COMPARATIVE PERFORMANCE ANALYSIS OF TRIPLE HYBRID FEATURE EXTRACTION TECHNIQUES COMBINED WITH VARIOUS CLASSIFIERS

| Feature Extraction Method (s) | Classifiers | Acc. (%) | Precis. (%) | Recall (%) | F1-Score (%) |
|---|---|---|---|---|---|
| DN-169 + LBP + HOG | SVM | 89.21 | 89.50 | 89.21 | 89.11 |
| DN-169 + LBP + HOG | XGBoost | 86.67 | 87.01 | 86.67 | 86.66 |
| DN-169 + LBP + HOG | LGBM | 89.01 | 89.01 | 88.47 | 88.46 |
| DN-169 + Gabor + HOG | SVM | 88.80 | 89.16 | 88.81 | 88.87 |
| DN-169 + Gabor + HOG | XGBoost | 86.74 | 87.01 | 86.74 | 86.78 |
| **DN-169 + Gabor + HOG** | **LGBM** | **90.12** | **90.66** | **90.23** | **90.16** |

Table 8. COMPARATIVE PERFORMANCE ANALYSIS OF HYBRID FEATURE EXTRACTION TECHNIQUES COMBINED WITH VARIOUS CLASSIFIERS

| Feature Extraction Method (s) | Classifiers | Acc. (%) | Precis. (%) | Recall (%) | F1-Score (%) |
|---|---|---|---|---|---|
| DN-169 + LBP + HOG + Gabor | SVM | 89.01 | 89.01 | 89.01 | 88.80 |
| DN-169 + LBP + HOG + Gabor | XGBoost | 90.04 | 90.16 | 90.01 | 90.01 |
| **DN-169 + LBP + HOG + Gabor** | **LGBM** | **92.88** | **93.01** | **92.42** | **92.45** |

## V. CONCLUSIONS

This paper presents a hybrid feature extraction framework that combines handcrafted features such as LBP, HOG and Gabor filters, along with deep features obtained from DenseNet-169 for PV panel defect classification. The experimental results ***(as illustrated in Fig. 6)*** show that the extensive feature fusion is inferior to selective feature fusion, and DenseNet-169 combined with Gabor has a good classification performance. The high-level semantic information is complemented with handcrafted visual information, improving the discriminative power of the framework. In contrast to the feature combinations in triple and quadruple features, these combinations had relatively poor performance, as they were increased in dimensionality, feature redundancy, and possible overfitting. The findings overall reveal the efficacy of the proposed framework for accurate and robust automated classification of PV panel defects, which shows great promise for practical application in monitoring PV panels.

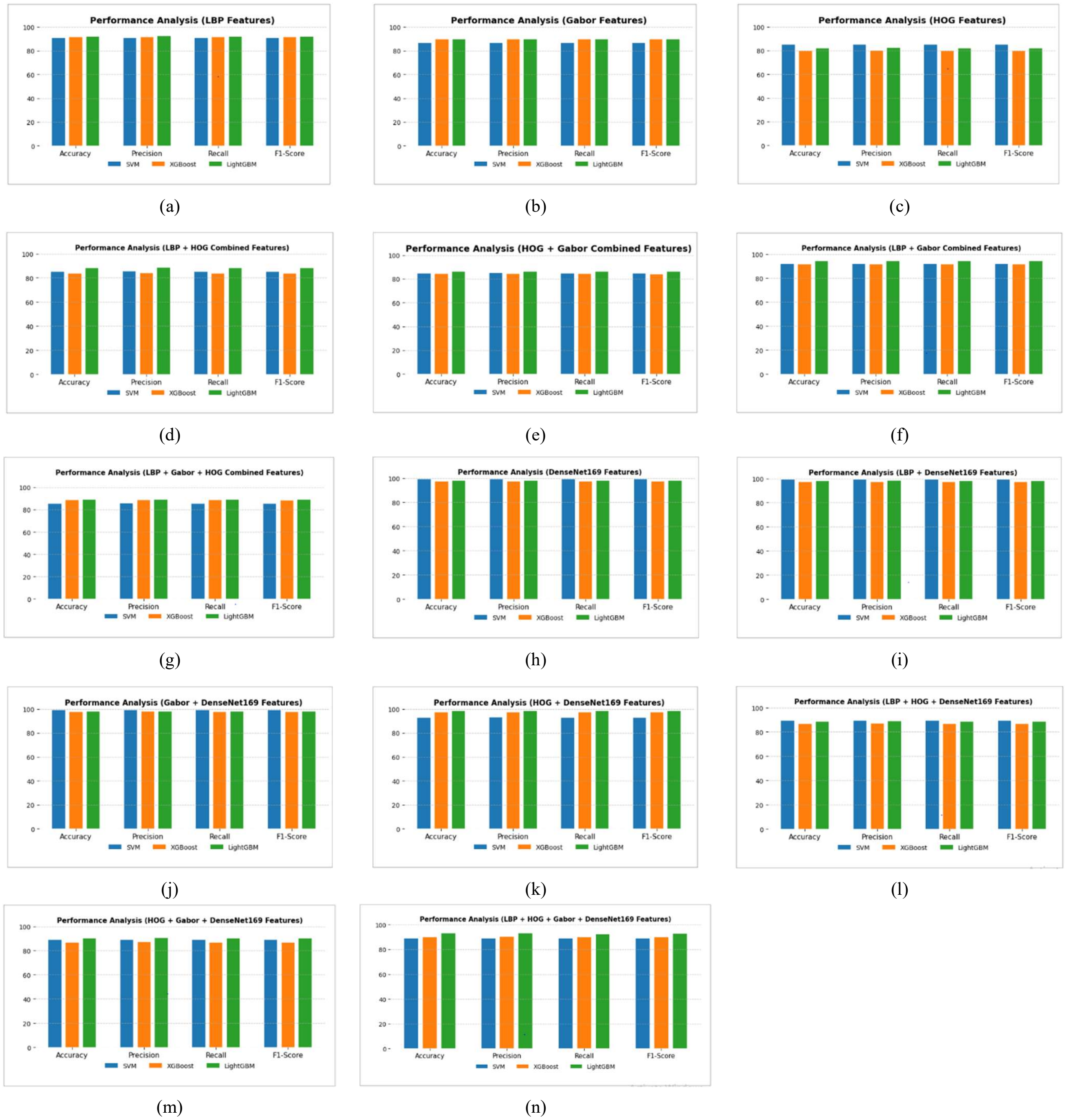

**Fig. 6.** Comprehensive performance comparison of SVM, XGBoost, and LightGBM models across a wide range of feature representations: (a) Local Binary Patterns (LBP), (b) Gabor texture features, (c) Histogram of Oriented Gradients (HoG), (d) LBP + HoG combined features, (e) HoG + Gabor combined features, (f) LBP + Gabor combined features, (g) LBP + HoG + Gabor combined features, (h) DenseNet169 deep features, (i) LBP + DenseNet169 combined features, (j) Gabor + DenseNet169 combined features, (k) HoG + DenseNet169 combined features, (l) LBP + HoG + DenseNet169 combined features, (m) Gabor + HoG + DenseNet169 combined features, and (n) LBP + HoG + Gabor + DenseNet169 fused features. Each bar chart presents Accuracy, Precision, Recall, and F1-Score, demonstrating the discriminative capability of individual and hybrid feature extraction techniques for rice grain classification.